\PassOptionsToPackage{prologue,dvipsnames}{xcolor}
\documentclass[sigconf]{acmart}
\usepackage[dvipsnames]{xcolor}

\AtBeginDocument{%
  }

\setcopyright{rightsretained}
\copyrightyear{2024}
\acmYear{2024}
\acmDOI{}

\acmConference[Conference KDDCup Workshop of SIGKDD '24]{KDDCup Workshop of the 30th ACM SIGKDD Conference on Knowledge Discovery and Data Mining}{August 25--29, 2024}{Barcelona, Spain}
\acmBooktitle{KDDCup Workshop of SIGKDD '24: KDDCup Workshop of the 30th ACM SIGKDD Conference on Knowledge Discovery and Data Mining, August 25--29, 2024, Barcelona, Spain}
\acmISBN{}

\usepackage{listings}
\usepackage{setspace}
\usepackage[normalem]{ulem}
\useunder{\uline}{\ul}{}

\newif\ifshowcomment
\showcommenttrue

\ifshowcomment
\newcommand{\yuanye}[1]{\textcolor{orange}{[Yuanye: #1]}}
\newcommand{\todo}[1]{\textcolor{red}{[TODO: #1]}}
\else
\newcommand{\yuanye}[1]{}
\newcommand{\todo}[1]{}
\fi

\begin{document}

\title{A Hybrid RAG System with Comprehensive Enhancement on Complex Reasoning}
\titlenote{We participated in Meta CRAG KDD Cup 2024 as Team ElectricSheep, securing third place in Task 1 and achieving first place in five of the seven question types in Task 2 among over $2,000$ participants and $5,500$ submissions. For access to the competition details and the leaderboard, please refer to the following URLs: \url{https://www.aicrowd.com/challenges/meta-comprehensive-rag-benchmark-kdd-cup-2024} and \url{https://discourse.aicrowd.com/t/meta-crag-challenge-2024-winners-announcement/10786}.}

\author{Ye Yuan}
\authornote{Ye Yuan is the leader of the team ElectricSheep.}
\affiliation{
\institution{State Key Laboratory of Multimedia Information Processing}
\institution{School of Computer Science}
\institution{PKU-Anker Embodied AI Lab}
\institution{Peking University}
\city{100871, Beijing}
\country{China}
}
\email{yuanye\_pku@pku.edu.cn}

\author{Chengwu Liu}
\affiliation{
\institution{State Key Laboratory of Multimedia Information Processing}
\institution{School of Computer Science}
\institution{PKU-Anker Embodied AI Lab}
\institution{Peking University}
\city{100871, Beijing}
\country{China}
}
\email{liuchengwu@pku.edu.cn}

\author{Jingyang Yuan}
\affiliation{
\institution{State Key Laboratory of Multimedia Information Processing}
\institution{School of Computer Science}
\institution{PKU-Anker Embodied AI Lab}
\institution{Peking University}
\city{100871, Beijing}
\country{China}
}
\email{yuanjy@pku.edu.cn}

\author{Gongbo Sun}
\authornote{The project was conducted as part of Gongbo Sun's research internship at the State Key Laboratory of Multimedia Information Processing, Peking University.}
\affiliation{
\institution{School of Computer, Data \& Information Sciences}
\institution{University of Wisconsin-Madison}
\city{Madison, WI}
\country{United States}
}
\email{gsun43@wisc.edu}

\author{Siqi Li}
\affiliation{
\institution{State Key Laboratory of Multimedia Information Processing}
\institution{School of Computer Science}
\institution{PKU-Anker Embodied AI Lab}
\institution{Peking University}
\city{100871, Beijing}
\country{China}
}
\email{xiaolilsq@stu.pku.edu.cn}

\author{Ming Zhang}
\authornote{Ming Zhang is the advisor of the team.}
\affiliation{
\institution{State Key Laboratory of Multimedia Information Processing}
\institution{School of Computer Science}
\institution{PKU-Anker Embodied AI Lab}
\institution{Peking University}
\city{100871, Beijing}
\country{China}
}
\email{mzhang\_cs@pku.edu.cn}

\renewcommand{\shortauthors}{Ye, et al.}

\begin{abstract}
Retrieval-augmented generation (RAG) is a framework enabling large language models (LLMs) to enhance their accuracy and reduce hallucinations by integrating external knowledge bases.
In this paper, we introduce a hybrid RAG system enhanced through a comprehensive suite of optimizations that significantly improve retrieval quality, augment reasoning capabilities, and refine numerical computation ability.
We refined the text chunks and tables in web pages, added attribute predictors to reduce hallucinations, conducted LLM Knowledge Extractor and Knowledge Graph Extractor, and finally built a reasoning strategy with all the references.
We evaluated our system on the CRAG dataset through the Meta CRAG KDD Cup 2024 Competition. Both the local and online evaluations demonstrate that our system significantly enhances complex reasoning capabilities. In local evaluations, we have significantly improved accuracy and reduced error rates compared to the baseline model, achieving a notable increase in scores. In the meanwhile, we have attained outstanding results in online assessments, demonstrating the performance and generalization capabilities of the proposed system.
The source code for our system is released in \textcolor{SkyBlue}{\url{https://gitlab.aicrowd.com/shizueyy/crag-new}}.
\end{abstract}

\ccsdesc[500]{Computing methodologies~Natural language generation}
\ccsdesc[500]{Computing methodologies~Information extraction}

\keywords{Large Language Models, Language Generation, Retrieval-Augmented Generation, Reasoning}


\maketitle

\section{Introduction}
Pre-trained large language models (LLMs), such as Llama3 \cite{llama3modelcard}, GPT-4 \cite{openai2024gpt4technicalreport}, Mistral-7B \cite{jiang2023mistral7b}, Gemini family \cite{geminiteam2024geminifamilyhighlycapable} and Qwen-2 \cite{yang2024qwen2technicalreport}, have demonstrated impressive advancements in the field of Natural Language Processing (NLP), notably in the Question-Answering task. This success is built on foundational knowledge, including factual knowledge \cite{petroni2019languagemodelsknowledgebases}, relational knowledge \cite{safavi2021relationalworldknowledgerepresentation}, and linguistic knowledge \cite{peters2018dissectingcontextualwordembeddings, goldberg2019assessingbertssyntacticabilities}, acquired by LLMs through exposure to the large scale internet corpus during training \cite{alkhamissi2022reviewlanguagemodelsknowledge}. Prior studies have shown the ability of LLMs in knowledge internalization and generation\cite{openai2024gpt4technicalreport, ouyang2022traininglanguagemodelsfollow, alkhamissi2022reviewlanguagemodelsknowledge,shen2024measuring,yuan2024measuring}, which knowledge is implicitly stored within model's parameters and retrieved during the generation process. 

These models typically undergo a two-stage training process: pre-training and post-training. Pre-training uses large-scale, task-agnostic data, and often needs trillions of tokens \cite{yang2024qwen2technicalreport, llama3modelcard, jiang2023mistral7b, openai2024gpt4technicalreport}, with the objective of next token prediction. However, models emerging from pre-training are not aligned with human values and can produce harmful, biased, or toxic content \cite{ouyang2022traininglanguagemodelsfollow}. To mitigate these issues, a post-training process is necessary, where techniques like supervised fine-tuning (SFT), Proximal Policy Optimization (PPO) \cite{schulman2017proximalpolicyoptimizationalgorithms} and Direct Preference Optimization (DPO) \cite{rafailov2024directpreferenceoptimizationlanguage} are applied \cite{llama3modelcard, yang2024qwen2technicalreport}.

Despite these advancements, under the paradigm of pre-training and fine-tuning, LLMs still face three critical challenges~\cite{yang2024cragcomprehensiverag, wu2024retrievalaugmentedgenerationnaturallanguage, huang2024survey}:

\textbf{1) Lack of domain expertise knowledge}: LLMs may struggle with specialized domains like laws and medicine due to limited exposure during pre-training. Internal knowledge along learned as parameters is insufficient to comprehensively address complex legal or medical issues \cite{long2024bailicaidomainoptimizedretrievalaugmentedgeneration}.  
Fine-tuning the model's weights in these areas requires external computational resources, which is expensive and time-consuming \cite{long2024bailicaidomainoptimizedretrievalaugmentedgeneration, ouyang2022traininglanguagemodelsfollow}. 

\textbf{2) Hallucination during generation}: Model can generate factually incorrect or inconsistent information \cite{openai2024gpt4technicalreport}. For instance, when asked ``Which number is larger, 3.11 or 3.9?'', most LLMs, including GPT-4, incorrectly respond with 3.11 > 3.9. The hallucination may accumulate progressively when the model incorporates prompting techniques like Chain of Thought (CoT) \cite{wei2023chainofthoughtpromptingelicitsreasoning}, a method used to augment LLMs reasoning ability \cite{vu2023freshllmsrefreshinglargelanguage}. 

\textbf{3) Difficulty integrating time-sensitive information}: LLMs often lack up-to-date information since the knowledge stored within the model parameters is static and does not undergo synchronous updates over time \cite{vu2023freshllmsrefreshinglargelanguage}, limiting their application in rapidly changing fields like sports and finance. These fields usually require real-time data processing such as querying the current price of a specific stock or the scores for a table tennis player at the Paris 2024 Olympics Games. However, previous research \cite{vu2023freshllmsrefreshinglargelanguage} indicates that LLMs like GPT-4 demonstrate an accuracy of less than 15\% when answering both slow and fast-changing questions. 

To address these issues, the Retrieval-Augmented Generation (RAG) approach has been introduced as a non-weight update approach that leverages external knowledge bases. A basic RAG system consists of two parts: a retriever and a generator. Relevant text documents are extracted from external knowledge sources and used as conditioning inputs alongside the query during generation  \cite{izacard2021leveragingpassageretrievalgenerative}. The retriever accurately computes the similarity between user queries and external factual knowledge using metrics such as cosine similarity. The top most relevant sentences are extracted from external databases and combined with the inputs for the generator. This process enables a general LLM to acquire domain-specific knowledge from a corresponding domain database without sacrificing its generalization capabilities. Additionally, by combining retrieved facts with input queries, the hallucination problem is mitigated, leading to more accurate and informed responses. Furthermore, by maintaining an up-to-date database, time-dependent information can be seamlessly integrated into the LLMs. Consequently, RAG provides an effective approach for enhancing the capability of LLMs to generate domain-specific, factual, and time-sensitive responses.

In this paper, we introduce a novel Retrieval-Augmented Generation (RAG) system designed for real-world applications. Our system was evaluated on the CRAG benchmark~\cite{yang2024cragcomprehensiverag}, achieving the third position in Task 1 and securing first place for $5$ out of $7$ question types in Task 2. The rest of the paper is structured as follows. In Section~\ref{sec:related-work}, we provide an overview of related work in the context of RAG system design. We introduce the CRAG benchmark in Section~\ref{sec:benchmark}. Our RAG system design is detailed in Section~\ref{sec:methods} and the complete system architecture is illustrated in Figure~\ref{fig:system-design}. The performance of our system is reported and discussed in Section~\ref{sec:experiments}. Finally, we conclude the paper in Section~\ref{sec:conclusion}, including a~\hyperref[sec:discuss]{Discussion Section} of potential improvements for future iterations of our system.

\section{Related Works}\label{sec:related-work}
Numerous techniques have been proposed to address the aforementioned issues.
For example, formal verification can help to reduce hallucination and output the verified reasoning process\cite{xiong2023trigo,wang2023dt,liu2023fimo}. Moreover, there are some efficient training methods\cite{shen2022palt,pan2023reusing-mango-nips,pan2024preparing-apollo-aaai} that can help the model adapt to domain-specific knowledge.
While numerous approaches have been proposed in the literature, the majority are tailored to address specific issues within a limited range of scenarios, making them unsuitable for direct application to CRAG tasks.
We present a compilation of recent research to highlight the diverse design choices within the broader RAG research community.
Drawing inspiration from recent advancements in the field, we have developed a novel design that integrates multiple strategies.

We built upon previous research \cite{huang2024survey, gao2023retrieval, wu2024retrievalaugmentedgenerationnaturallanguage,yuan2024measuring} to structure the conventional RAG workflow into four phases: pre-retrieval, retrieval, post-retrieval, and generation.

\subsection{Pre-retrieval}

The pre-retrieval phase involves indexing, query manipulation, and data modification.
That is, compiling and indexing external knowledge sources to enhance the efficiency of subsequent searching execution, refining the query to better match the external data distribution and modifying data sources to create a coherent representation that encapsulates relevant knowledge for further reasoning.

\noindent\textbf{Indexing:}
Text indexing has emerged as a prominent area of research within the field of data mining, leading to the development of numerous methodologies.
Traditional indexing methods include the inverted index \cite{zobel2006inverted}, suffix tree \cite{mccreight1976space}, and term frequency-inverse document frequency (TF-IDF) index \cite{sparck1972statistical}.
These indexing methods utilize term frequency so that they may neglect the semantic information inherent in the text.
Recent indexing methods employ language models such as BERT \cite{devlin2018bert} and Sentence-T5 \cite{ni2021sentence} to encode text into vectors within latent space.
These vectors are used to construct vector indexes \cite{douze2024faiss} that enhance similarity search through techniques including product quantization (PQ) \cite{jegou2010product}, Hierarchical Navigable Small World (HNSW) \cite{malkov2018efficient}, and Navigating Spreading-out Graph (NSG) \cite{fu2017fast}.

\noindent\textbf{Query Manipulation:} The manipulation algorithms discussed below contribute to developing an improved metric for evaluating the similarity between user queries and external data.
It is desirable to formulate a fully specified, context-independent query that effectively articulates user intent and aligns with the distribution of external knowledge sources.
Early research primarily utilizes an n-gram language model derived from hundreds of millions of web search sessions \cite{jansen2009patterns} or employs a pointer-generator network to improve referring expression resolution.
Recent studies have leveraged the few-shot or zero-shot learning capabilities of language models that utilize a small set of manually created rewriting labels \cite{yu2020few, yu2022generate}.
In contrast to traditional query rewriting, which follows a ``query in, query out'' format, recent query manipulation approaches utilizing LLMs allow for ``query in, passage out'' approaches \cite{yu2022generate, ma2023query}.

\noindent\textbf{Data Modification:} Data extracted from original external knowledge sources may contain elements that are not conducive to the reasoning process of LLMs, such as redundant HTML tags found in web pages.
However, these processes are closely linked to the format and distribution of external knowledge sources and often involve tedious text normalization procedures that are largely unrelated to the reasoning process, including tokenization, stemming, and the removal of stop words.
Prior research has examined the use of substrings as document identifiers to improve search capabilities \cite{bevilacqua2022autoregressive}, as well as the application of LLMs to recite one or more relevant passages from their internal memory \cite{sunrecitation}.

\subsection{Retrieval}

The retrieval phase primarily involves searching for pertinent materials from external knowledge sources and ranking them according to their relevance to the user query.

\noindent\textbf{Search:} Searching for relevant documents efficiently from vast external knowledge sources, such as the entire Internet, can be exceedingly challenging.
Previous research on RAG that aims to retrieve local data typically employs local vector databases, such as Elasticsearch \cite{gormley2015elasticsearch}, or utilizes libraries like FAISS \cite{douze2024faiss} for efficient similarity search and vector clustering.
On the other hand, research focused on general-purpose RAG typically utilizes external web search engines directly; for instance, WebGPT \cite{nakano2021webgpt} employs the Bing API.

\noindent\textbf{Ranking:} The ranking metric establishes the priority of each document housed in external knowledge sources.
The metric used to evaluate the similarity between a potentially manipulated query and a document can be classified into two categories: sparse retrieval and dense retrieval. 
Sparse retrieval metrics, such as BM25 \cite{robertson2009probabilistic, izacard2021leveragingpassageretrievalgenerative} and Term Frequency–Inverse Document Frequency (TF-IDF) \cite{chen2017reading}, rely on word frequency, whereas dense retrieval metrics utilize embedded vector, including Euclidean distance \cite{khandelwal2019generalization}, cosine similarity \cite{juvekar2024cos}, and Dense Passage Retrieval (DPR) \cite{karpukhin2020dense, izacard2021leveragingpassageretrievalgenerative}.

\subsection{Post-retrieval}

The post-retrieval phase includes re-ranking and filtering, which further refines the ranking results, and filtering out materials that are irrelevant to the querying topic.

\noindent\textbf{Re-ranking:} Although vector similarity-based document retrieval can be executed efficiently and effectively, including in parallel \cite{johnson2019billion, khattab2020colbert} or distributed systems \cite{echihabi2021new}, its capacity to reveal semantic relationships between queries and documents remains constrained \cite{steck2024cosine}.
This is where re-ranking becomes significant: typically bigger models, which offer greater accuracy but lower efficiency, can precisely reorder the limited set of documents retrieved.
Some approaches \cite{glass2022re2g, ram2023context} embed the query-document pair in a single pass, facilitating cross-attention between sequences.
These approaches enhance the evaluation of mutual information between the texts.
Other studies employ large language models as few-shot annotators to generate data for training a cross-attention re-ranker \cite{dai2022promptagator}, or investigate the auto-regressive generation of re-ranking results to leverage inter-document relationships \cite{hofstatter2023fid}.

\noindent\textbf{Filtering:} Filtering seeks to remove redundant parts from the retrieval results, which may either be of poor quality or exhibit low relevance to the user query \cite{wang2023learning}.
Current studies employ large language models to evaluate the utility of retrieved information and to critique whether all verification-worthy statements are substantiated by the associated documents \cite{asai2023self, wang2024blendfilter}, or condense retrieved documents into a textual summary, thereby minimizing inference costs \cite{xu2023recomp}.

\subsection{Generation}

The curated reference materials are subsequently processed in the generation phase to produce the final results.
Additionally, various enhancement techniques and customizations can be applied to the same phrase.

\noindent\textbf{Enhancing:} Enhancing, also known as retrieval fusion, seeks to improve the performance of large language model generations by utilizing retrieved documents.
Typically, the retrieved documents are concatenated with the query in the expectation that the language models will generate coherent results based on the reference materials \cite{lewis2020retrieval, ram2023context, guu2020retrieval}.
Other methods utilize sophisticated prompt templates that integrate multiple pieces of information \cite{vu2023freshllmsrefreshinglargelanguage} or employ techniques such as context compression, summarization, and filtering to achieve more efficient inference \cite{arefeen2024leancontext, liu2023tcra, wang2023learning, xu2023recomp}.
Several studies have also investigated the integration of encoded retrieval vectors into input features \cite{izacard2021leveragingpassageretrievalgenerative, izacard2023atlas, singh2021end}.

\noindent\textbf{Customization:} During the customization process, it is crucial to meticulously refine the model's outputs to ensure alignment between the model's responses and the user's intended queries.
This refinement will result in final answers that are concise, informative, and well-formatted.
These approaches include generating reflection tokens for self-evaluation of the model's output \cite{asai2023self} and implementing a graph-text contrastive learning method to enhance the alignment between generated content and retrieved references\cite{kang2023knowledge}.

\section{CRAG Benchmark}\label{sec:benchmark}
CRAG benchmark~\cite{yang2024cragcomprehensiverag} is a factual question-answering benchmark with thousands of QA pairs and 50 real-world web pages for each data. The benchmark also provides a mock API for Knowledge Graph(KG) searching. The benchmark is set for the KDD Cup 2024 with $2,706$ data items available in public, half of which are for public validation. There are 5 domains and 8 question types in the benchmark, and each data item has a \verb|static_or_dynamic| label that indicates whether the answer to a question changes and the expected rate of change, which can be used to analyze the model strengths and weaknesses. In order to mimic real-world application scenarios, each generated response is limited to 30 seconds on an AWS G4dn.12xlarge instance, which is equipped with 4 NVIDIA T4 GPUs providing a total of 64 GB of GPU memory during inference.

The benchmark is split into three tasks in the competition: Retrieval Summarization, Knowledge Graph and Web Retrieval, and End-to-End Retrieval Augmented Generation, which are from simple to complex. We introduce the three tasks as follows:
\begin{itemize}
    \item \textbf{Task 1: Retrieval Summarization.}
    In this task, each question is provided with 5 web pages potentially containing relevant information to mimic the top 5 results from a real-world web search. They are created by storing up to 50 pages from search queries related to each question. Web pages are quite long, containing about $120,000$ tokens on average, to measure the CRAG systems' capability to identify and condense this information into accurate answers.
    \item \textbf{Task 2: Knowledge Graph and Web Retrieval.}
    Besides the web pages provided with each question, this task introduces mock APIs to access information from underlying mock Knowledge Graphs (KGs), with structured data possibly related to the questions. 
    Mock KGs are created using the data behind the questions, supplemented with "hard negative" data to simulate a more challenging retrieval environment. Mock APIs facilitate structured searches within these KGs and can be used with parameters derived from the questions, to retrieve relevant data for answer formulation. The evaluation focuses on the systems' ability to query structured data and integrate information from various sources into comprehensive answers.
    
    \item \textbf{Task 3: End-to-End Retrieval Augmented Generation.}
    The third task increases complexity by providing 50 web pages and mock API access for each question, encountering both relevant information and noises. It assesses the systems' skill in selecting the most important data from a larger set, reflecting the challenges of real-world information retrieval and integration.
\end{itemize}

Each task builds upon the previous, employing more demanding tasks to gauge the CRAG system's capabilities in real-world and intricate scenarios.

As for evaluation, CRAG Benchmark employs both automated (Auto-eval) and human (Human-eval) evaluations. Auto-eval employs rule-based matching and GPT-4 assessment to check answer correctness. To promote concise answers, Auto-evaluators will only consider responses within $50$ tokens. Human annotators will only decide the rating of top teams to reduce workload. In addition to correctness, human eval requires basic fluency for an answer to be considered Perfect.

Notably, the competition evaluates the system with a positive score for the correct answer, a zero score for answering ``I don't know'', and a negative score for the wrong answer, which is noted as \verb|hallucination|. The benchmark not only encourages the model to answer the question correctly but also punishes the model for incorrect responses, which is closer to the real-world scenario.

\section{System Design}\label{sec:methods}
\begin{figure*}[t]
    \centering
    \includegraphics[width=0.95\linewidth]{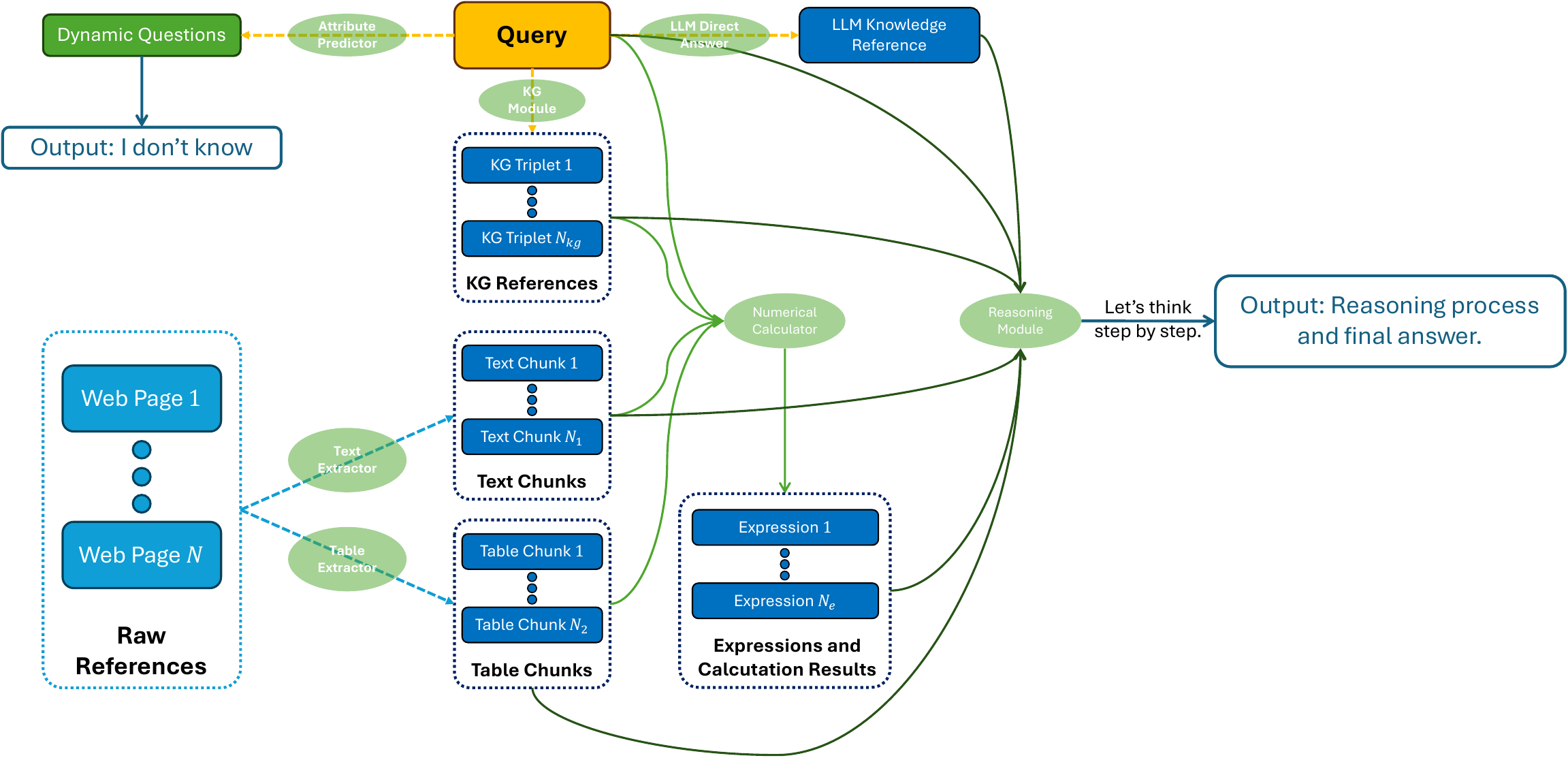}
    \caption{The complete design of our system. There are two possible routes for the generation. If the query is classified by the in-context learning as ``dynamic'', we will output ``I don't know'' directly to reduce hallucination on these hard problems.}
    \label{fig:system-design}
    \vspace{-0.5cm}
\end{figure*}

The complete design of our system is shown in Figure~\ref{fig:system-design}. There are $6$ critical modules in our system, including (1) web page processing, (2) attribute predictor, (3) numerical calculator, (4) LLM knowledge extractor, (5) KG Module, and (6) reasoning module. We have enhanced the system's capabilities in information extraction, reducing hallucinations, numerical calculation accuracy, and higher-order reasoning through these modules. Additionally, we have implemented special handling for corner cases. We will introduce these modules as follows.

\subsection{Web Page Processing}

Web pages serve as a shared information source for all three tasks, containing a substantial amount of potentially valuable information that can aid the model in task completion.
As a result, web page processing is a critical component of system design, directly impacting both the quality of the extracted information and the accuracy of subsequent language model generations.

\begin{figure}[ht]
    \centering
    \includegraphics[width=0.95\linewidth]{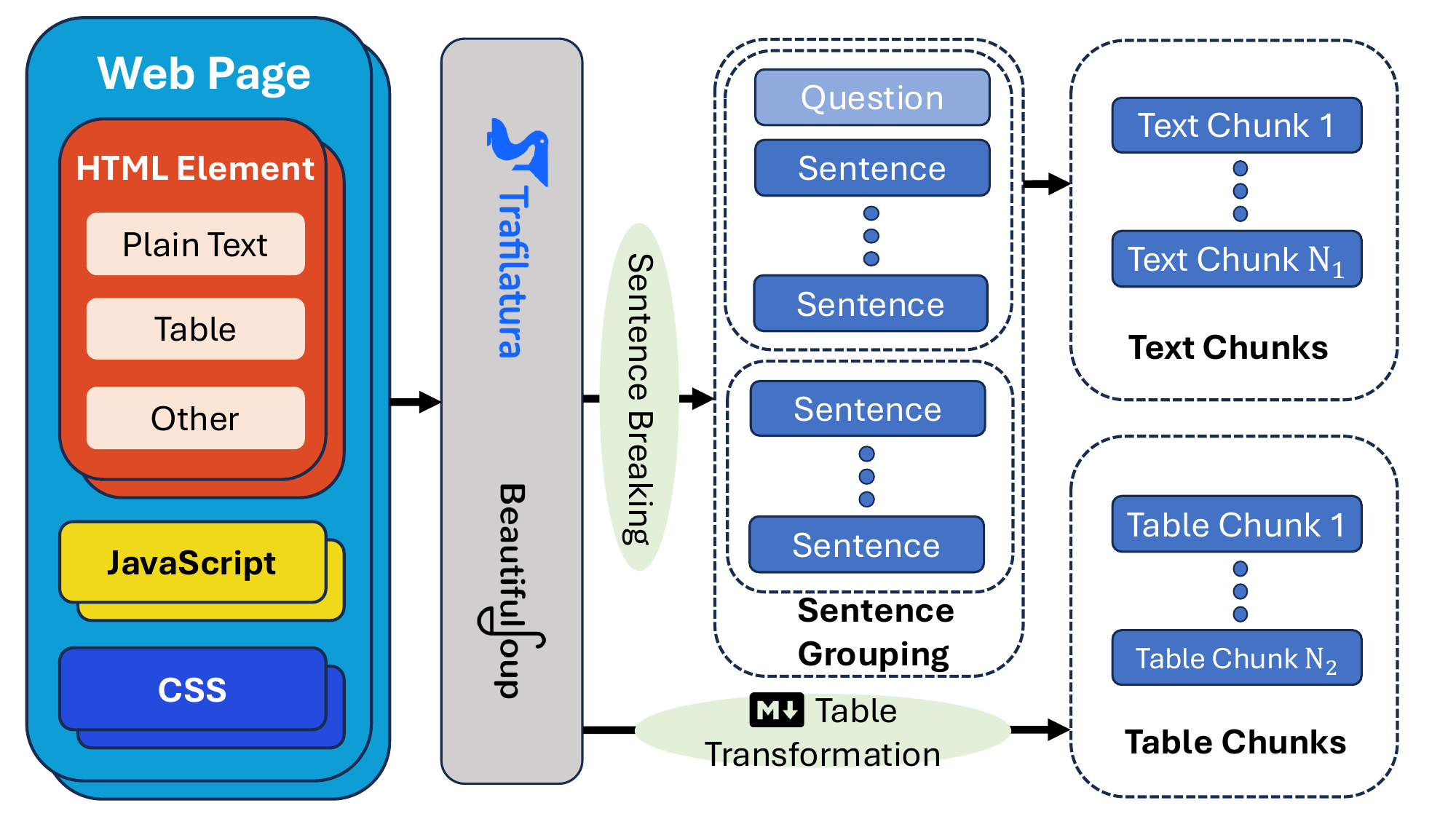}
    \caption{The design of our web page processing. We utilized Trafilatura and BeautifulSoup to extract plain text and tables from web pages. Following this extraction, we employed Blingfire to segment the plain text into sentences, which were then grouped into chunks based on heuristic methods. Additionally, the tables were converted into Markdown format for further processing.}
    \label{fig:web_page_processing}
    \vspace{-0.5cm}
\end{figure}

However, despite the abundance of information presented in natural language on web pages, extracting this information is not straightforward.
This complexity is due to the frequent presence of significant amounts of noise that does not contribute relevant information necessary for task completion.
Such noise can unnecessarily prolong the model's processing and reasoning time, potentially leading to misinterpretations.
The types of noise encountered include decorative HTML tags used for typography, JavaScript code, and internal comments within the web page. While some HTML tags may contain semantic information that aids in paragraph segmentation or title identification, the useful information is not easy to extract. Moreover, there is some structured information in the HTML like tables, which will do harm to the text quality if they are improperly handled through methods such as truncation or splicing other texts. So we process the raw web page into two parts: text chunks and tables, to get the references with higher quality.

\noindent\textbf{Text Chunks Processing.} To address the challenges posed by the complexity of modern HTML web pages, we adopt \verb|trafilatura|, a Python library specifically designed for gathering text from the Web.
This library effectively mitigates noise generated by recurring elements such as headers, footers, and links. We also clean all the tables by identifying the \verb|<table>| tag, which will be processed specially.
To enhance robustness, we employ the classic \verb|BeautifulSoup| library as a fallback option for a small subset of web pages that \verb|trafilatura| cannot process.
After extracting text using these two tools, we utilize functions provided by the \verb|Blingfire| library to segment the text into individual sentences.

The meaning of a sentence is often influenced by its contextual placement; therefore, the information within a single sentence is often incomplete.
We organize sentences into chunks based on the following rules to enhance semantic coherence in subsequent retrieval processes.
First, we truncate individual sentences that exceed a predetermined length threshold to ensure they remain within an appropriate length.
Next, we implement a keyword-based approach to identify whether a sentence is a question, utilizing indicators such as the 5W1H interrogatives at the beginning and the presence of question marks at the end.
All keywords utilized in this process can be found in Appendix~\ref{app:ques-classify}.
Sample data analysis has shown that questions are often immediately followed by their corresponding answers.
Therefore, we concatenate questions with their subsequent text until we reach the pre-assigned length threshold.
Finally, we connect any remaining ungrouped sentences in sets of $3$.

\noindent\textbf{Tables Processing.} Given that the extracted text data typically does not include tables, we have employed \verb|BeautifulSoup| to extract tables from web pages and convert them into Markdown format.
We hypothesize that exposure to numerous documents formatted in Markdown during the model training phase will improve the model's understanding and interpretation of this format. Finally, we cleaned the empty tables to reduce the noise. The source code for table transformation is listed in Appendix~\ref{app:table-transformation}.

\noindent\textbf{Text Embedding \& Ranking Metrics.}
Regarding the ranking metrics and methodologies for retrieval, it is important to note that in the CRAG tasks, each question is associated with a relatively small number of candidate articles (five for each question in the initial two tasks).
Consequently, we have adopted a straightforward approach.
We utilized the \verb|sentence-t5-large| model \cite{ni2021sentence} to generate vector embeddings for both the text chunks and the queries, and we employed cosine similarity as the metric to rank the relevance of the text chunks.
The cosine similarity between the user query embedding and a text chunk embedding \( \mathbf{q} \) and \( \mathbf{c} \) is defined as the cosine of the angle between them. Mathematically, it is given by:

\[
\text{cosine\_similarity} = \cos(\theta) = \frac{\mathbf{q} \cdot \mathbf{c}}{\|\mathbf{q}\| \|\mathbf{c}\|}
\]
where \( \mathbf{q} \cdot \mathbf{c} \) is the dot product of embedding vectors \( \mathbf{q} \) and \( \mathbf{c} \), and \( \|\mathbf{q}\| \) and \( \|\mathbf{c}\| \) are their respective magnitudes.
Our experiments indicated that employing LLMs for complex query manipulation methods did not significantly improve retrieval accuracy; instead, it resulted in substantial computational overhead.
Therefore, we primarily leveraged the LLM knowledge extractor, as detailed below, to enhance the quality of the references.

\subsection{Attribute Predictor}

Large language models demonstrate considerable variability in their performance across various question-answering tasks.
Within the context of the CRAG tasks, answering aggregation questions and multi-hop questions poses greater challenges than addressing simple questions. This difficulty arises from the model's need to not only possess robust information retrieval capabilities but also to integrate multiple sources of information and engage in reasoning processes.
Moreover, when dealing with questions related to slow-changing and fast-changing facts, the model must exhibit temporal awareness, which adds complexity to the generation of accurate responses \cite{vu2023freshllmsrefreshinglargelanguage}.
To tackle these challenges, we have developed an attribute predictor that assesses the type of each specific question and the rate of underlying factual change, aiming to optimize performance across all question types.

There are three useful attributes in the benchmark: \verb|domain|, \verb|question_type|, and \verb|static_or_dynamic|.
Following the descriptions outlined in the CRAG dataset, we classified the  \verb|domain| into five categories: Finance, Sports, Music, Movies, and Encyclopedia Open Domain. The \verb|question_type| was divided into the following categories: simple question, simple question with some condition, set question, comparison question, aggregation question, multi-hop question, post-processing question, and false premise question. However, the \verb|question_type| attribute is hard to predict because it often needs reasoning through all references, so we didn't apply attribute prediction to it.
Additionally, the \verb|static_or_dynamic| attribute, which pertains to timeliness, was classified into four categories: real-time, fast-changing, slow-changing, and stable. Moreover, we found that the boundaries between these four categories are not clear, so we only classified this attribute into two categories: static and dynamic. For each attribute, we implemented two methods for question classification: one leveraging the in-context learning capability of large language models and the other employing support vector machines (SVM).

\noindent\textbf{In-Context Learning.} Large language models demonstrate robust natural language understanding and strong multi-task generalization abilities.
We prompt the model with classification instructions and $5$ demonstrations for categories, instructing it to classify subsequent questions.
The demonstrations are randomly selected from the publicly available CRAG dataset.
All prompts utilized in the classification process can be found in Appendix~\ref{app:prompt-attr-predictor}.

To enhance classification reliability, we adopted a self-consistency strategy involving multiple samplings from the large language model. The category that appeared most frequently among the sampled results was designated as the classification for the question.
\noindent\textbf{SVM.} We also tried to train an SVM classifier using the CRAG public dataset to reduce computational overhead. We used the \verb|all-MiniLM-L6-v2| model to get the sentence embeddings, which are used to train the SVM.
We observed that the SVM model can get higher accuracy in predicting the attributes with less time and computation consumption.
However, we didn't have time to merge the code into our final version for evaluation.
So, consequently, the submitted version relied on the few-shot learning approach with the large language model for classification, and we leave the improvement for this module as future work. We will show the detailed results analysis and comparison in Section~\ref{sec:analysis}.

\subsection{Numerical Calculator}

Previous research has highlighted the phenomenon of ``hallucination'' in large language models, particularly concerning their performance in precise numerical calculations \cite{polu2020generative, geva2020injecting}.
Within the framework of the CRAG tasks, answers to aggregation questions, multi-hop questions, and post-processing questions are not directly available in the retrieved content.
Instead, these tasks require the model to derive the final answer through a series of reasoning steps, which often involve precise numerical computations, especially in the domain of finance and sports.
This requirement poses additional challenges to the model's capacity to generate accurate results.

To address this issue, we employed an approach that leverages external tools, drawing inspiration from previous research \cite{schick2023toolformer,paranjape2023art,yuan2024measuring}.
We encourage the large language model to articulate the reasoning steps necessary to solve the problem as mathematical expressions while delegating the actual numerical calculations to an external Python interpreter.
We specifically integrate retrieved text chunks and tables that may contain numerical information into the model's prompts and employ prompt techniques that encourage the model to generate valid Python expressions directly.
Detailed specifications of the prompts are provided in the Appendix~\ref{app:prompt-calculator}
In the submitted version, we utilized multiple sampling and processed the generated Python expressions using the \verb|eval| function.

Note that the program code generated by the LLM may include malicious code, and executing such code directly poses a potential threat to system stability.
To mitigate such risk, the best practice for ensuring system security is to use \verb|ast.literal_eval| or to execute the code within a sandbox environment. 
We leave the safety and proper termination of sandbox execution as future work.

\subsection{LLM Knowledge Extractor}

Through extensive training on diverse corpora, LLMs have acquired substantial knowledge and demonstrated robust question-answering capabilities across a wide range of domains.
Chances are that the reference documents are not beneficial in generating accurate responses to the model's queries.
This may be attributed to the possibility that the retrieved materials are outdated, or they may include irrelevant or even misleading information.
Consequently, these reference documents do not enhance the model's capacity to produce answers but harm it.
Furthermore, during the generation phase, we employ the Llama 3 instruction-tuned models that are optimized for dialogue use cases and possess a robust capability for following instructions.
Our findings indicate that when reference documents are provided within the prompt, the model exhibits a significant tendency to extract answers from these documents, regardless of whether the documents contain the necessary information.
This inclination may be a result of the instructional fine-tuning process.
In contrast, LLMs that operate without any references are capable of accurately answering these questions.
In light of this observation and drawing inspiration from prior research \cite{sunrecitation}, we developed a large language model knowledge extractor.
This extractor leverages the knowledge-rich responses generated by the large language model as part of the reference materials for enhanced reasoning.

The process of extracting knowledge from the model closely resembles the normal model generation process.
It similarly utilizes zero-shot indications, which include prompts requiring the model to assess whether a given query pertains to a false-premise issue and to generate more concise responses.
However, a notable distinction exists in the lack of reference documents sourced from external knowledge bases within the prompts, as well as the exclusion of multiple sampling, which is intended to reduce computational overhead.
In this way, the LLM could respond solely based on the knowledge internalized within its parameters during the training.
Our findings suggest that this approach results in favorable performance on questions classified as slow-changing and stable.
We anticipate that this approach will effectively align the knowledge embedded in the LLM's parameters with external reference documents, thereby mitigating the issue of the model's excessive dependence on externally retrieved information. Moreover, we also use the zero-shot CoT to let the model reasoning by itself for more accurate knowledge. The prompt template is shown in Appendix~\ref{app:prompt-knowledge-extractor}

However, as described previously, letting the model directly answer the questions will introduce hallucinations in its knowledge, although with zero-shot CoT reasoning. To balance the hallucination and knowledge from the LLM itself, we only treat the output of this module as one of the references. We have carefully designed prompts to ensure that the model neither overly relies on document references nor excessively trusts the LLM's knowledge. Section~\ref{sec:reasoning} will provide a more detailed introduction.

\subsection{Knowledge Graph Module}
In addition to web references, Task 2 and Task 3 also provide a mock API for querying the provided knowledge graph (KG). As a structured knowledge base, a KG provides accurate information. However, the generation of a KG query is crucial to determining whether the system can retrieve the correct answer. We started from the KG baseline, which extracted the entities in the query with an LLM, and generated the query by manual rules. The quality of rule-based queries is limited by the complexity of the rules, and hard to scale. So we tried a function-calling method, which makes all the mock APIs as the input of the LLM, and lets it generate a proper function calling. However, due to limitations in time and resources, we were unable to optimize the models and prompts for the function-calling method, resulting in suboptimal performance. Therefore, we reverted to the KG baseline method and did not make further improvements in the submitted version. We show the prompts for the function-calling methods in Appendix~\ref{app:prompt-kg}.

\subsection{Reasoning Module}
After all the previously introduced processing methods, we get text chunks, tables, triplets from KG, and knowledge from LLM weights as the references. We carefully designed a prompt template to let the LLM do reasoning from all these references and get the final answer. We control the reasoning process by output format demonstration and zero-shot CoT, which is useful for multi-hop questions. Leveraging the strong instruction-following capabilities of Llama3-70B-Instruct, we've successfully maintained steady progress in controlling reasoning tasks. We designed several rules to constrain the reasoning path and output format, including that the output should be precise, and guide the model reasoning by asking intermediate questions in the prompt. The complete prompt is shown in Appendix~\ref{app:prompt-reasoning}.
\label{sec:reasoning}
\subsection{Handling Corner Cases}
In addition to the main modules mentioned above, we have also handled many corner cases, including (1) identifying invalid questions; (2) encouraging the model to answer ``I don't know'' for unsure answers to reduce hallucination; and (3) dealing with outputs that do not comply with the instruction format. We will introduce our design to handle these corner cases as follows.

\noindent\textbf{Invalid Questions.}
There are some questions that have false premises, which means the query is contradictory to the fact. For these questions, the model should output ``invalid questions''. To identify this type of question, the model needs to carefully analyze the references provided. We add special rules in the reasoning prompt shown in Appendix~\ref{app:prompt-reasoning}

\noindent\textbf{Reduce Hallucination.}
We employed two approaches to alleviate hallucinations: attribute prediction and reasoning. We found that the time-changing questions, which would be labeled as \verb|dynamic| by attribute predictor, are hard for our system and we do not have enough time and resources to improve them. So we manually let the system answer ``I don't know'' for these questions. Moreover, we added several rules and prompt engineering techniques in the reasoning module to let the model answer ``I don't know'' when it is unsure. Ultimately, we configured the system to exclusively output "I don't know" and refrain from adding any additional words whenever "I don't know" is included in the initial response.

\noindent\textbf{Incorrect Format.}
Cause we didn't conduct constrained sampling for the reasoning output, there is the possibility that the model will output answers that can not be parsed. To handle this situation, we design a backup summarization agent to summarize the final answer precisely and concisely based on the reasoning module's output when the parse fails.
The prompt for this module is shown in Appendix~\ref{app:prompt-backup}

\begin{table}[!t]
\resizebox{0.45\textwidth}{!}{
\begin{tabular}{@{}l|ccc|c@{}}
\toprule
                      & Correct(\%)$\uparrow$   & Missing(\%)   & Hallucination(\%)$\downarrow$ & Score(\%)$\uparrow$     \\ \midrule
Official LLM Baseline & 16.2          & 0.0           & 83.7              & -67.6         \\
Official RAG Baseline & 25.4          & 2.5           & 72.1              & -46.6         \\
\textbf{Our System}   & \textbf{29.7} & \textbf{56.3} & \textbf{13.9}     & \textbf{15.8} \\ \bottomrule
\end{tabular}
}
\caption{The main results of our designed system evaluated in the public test dataset.}
\label{tab:main-result}
\vspace{-1cm}
\end{table}

\section{Experiments}
\label{sec:experiments}
We conducted ample experiments to verify the effectiveness of each module. The main results of our local evaluation are shown in Table~\ref{tab:main-result}, where we got a lot of improvement in the public test set compared with the baseline of Task 1. We got a $15.8\%$ score where we greatly reduced the hallucination ratio and changed these hallucinations into answering ``I don't know''. We show detailed analysis and ablation studies of our evaluation results. In the final private test, we got a $21.8\%$ score in Task 1. We will also show an analysis of the private evaluation. We will use ``Official RAG Baseline'' as the baseline model of Task 1 in the following tables.

\subsection{Detailed Analysis}
\label{sec:analysis}
\begin{figure}[ht]
    \centering
    \includegraphics[width=0.95\linewidth]{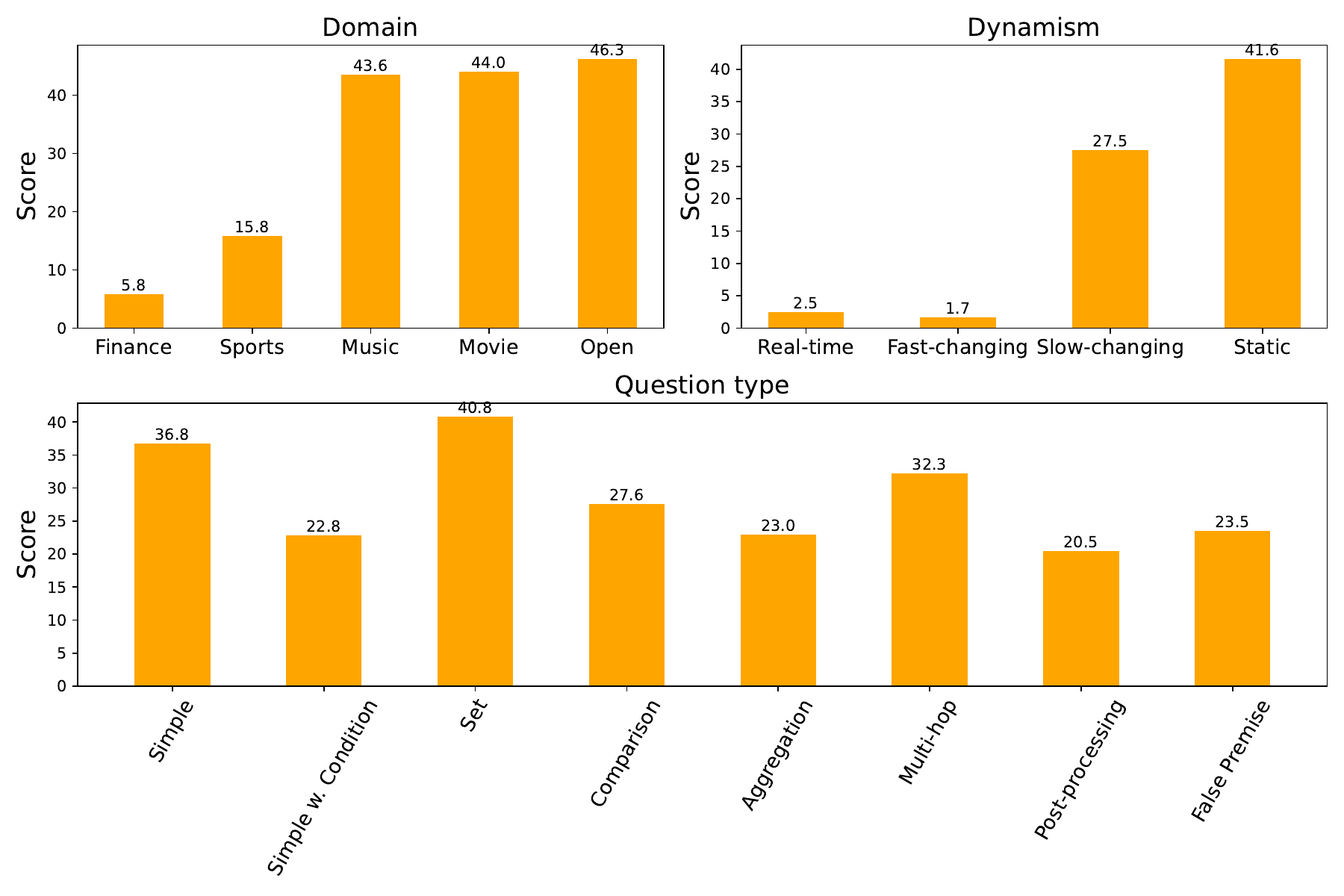}
    \caption{Detailed score across different attributes in the local evaluation of Task 1.}
    \label{fig:detailed-analysis}
\end{figure}

To gain a deeper understanding of the strengths and weaknesses of our system across various aspects, we conducted a meticulous analysis of the evaluation results on the public test set. Figure~\ref{fig:detailed-analysis} shows the detailed scores in Task 1 setting.

For the domain attribute, we perform well in areas such as movies, music, and open topics, but our performance is lacking in finance and sports. This is due to the fact that these two domains require the model to have the ability to answer dynamic information that changes over time. To prevent hallucinations, our model opts to refuse to answer such queries. The performance regarding the attribute of dynamism reaches the same conclusion: as the dynamism of the model increases, the effectiveness of our system gradually declines. Regarding the score distribution across question types, our analysis indicates that the system exhibits superior performance on tasks requiring complex reasoning, which benefits from the robust functionality of our integrated reasoning module.

\subsection{Ablation Study}
We performed extensive experiments to validate the enhancements contributed by individual components of our system. We developed the system incrementally, systematically verifying and integrating beneficial modules. Consequently, comprehensive ablation studies, which would involve removing each component from the final system, were not conducted. Instead, we documented the rationale behind the inclusion of each module and its resulting improvements. Table~\ref{tab:ablation} outlines the primary construction pathway of our system.

We started from the baseline model of Task 1\footnote{The model can be found {\color{magenta}\href{https://gitlab.aicrowd.com/aicrowd/challenges/meta-comprehensive-rag-benchmark-kdd-cup-2024/meta-comphrehensive-rag-benchmark-starter-kit/-/blob/master/models/rag_llama_baseline.py}{here}}.} and did a lot of refinement and added modules to it. As shown in Table~\ref{tab:ablation}, each module we added would increase the final score. The main optimization directions are reducing hallucinations and increasing correctness.

\begin{table}[]

\resizebox{0.45\textwidth}{!}{
\begin{tabular}{@{}l|ccc|c@{}}
\toprule
                                                                                & Correct (\%)$\uparrow$ & Missing (\%)  & Hallucination (\%)$\downarrow$ & Score$\uparrow$     \\ \midrule
Official RAG Baseline                                                           & 16.2                   & 0.0           & 83.8                           & -67.6               \\
+ Optimized Chunk Extraction                                                    & 23.7                   & 0.0           & 76.3                           & -52.6               \\
\begin{tabular}[c]{@{}l@{}}+ Llama3-70B-GPTQ\\ \ \ \ \& Prompt Refinement\end{tabular} & 24.4                   & 50.8          & 24.8                           & -0.3                \\
+ Attribute Predictor                                                           & 19.9                   & 65.4          & 14.7                           & 5.2                 \\
+ Table Extractor                                                               & 21.3                   & 63.4          & 15.2                           & 6.1                 \\
+ Numerical Calculator                                                          & 23.7                   & 29.7          & 16.6                           & 7.2                 \\
+ Reasoning Module                                                              & 20.9                   & 67.3          & 11.8                           & 9.1                 \\
+ False Premise Identify                                                        & 26.0                   & 60.5          & 13.5                           & 12.5                \\
+ LLM Knowledge Extractor                                                       & 27.3                   & 59.3          & 13.4                           & 13.9                \\ \midrule
\textbf{\begin{tabular}[c]{@{}l@{}}+ KG Module\\ (Final System)\end{tabular}}   & \textbf{29.7}          & \textbf{56.3} & \textbf{13.9}                  & {\ul \textbf{15.8}} \\ \bottomrule
\end{tabular}
}

\caption{Ablation results of our system. The modules are gradually added to the system. The results are tested in Task 1 and Task 2, which are only different in using KG.}
\label{tab:ablation}
\vspace{-0.5cm}
\end{table}

\subsection{Analysis For Private Evaluation}
Although the competition organizers have not provided a comprehensive and detailed analysis of the results on the private leaderboard, we can still present the currently published results and analyze our system. Table~\ref{tab:private-main} shows the score for Task 1 in private evaluation. Our system achieved scores close to the champion in Task 1 but fell significantly behind in Task 2 and Task 3. We believe this is due to our underutilization of the information from the knowledge graph. Table~\ref{tab:task-2-prizes} shows the prizes we won from 5 out of 7 question types in Task 2. We find that our system performs well on question types that require complex reasoning, such as aggregation and multi-hop, which we attribute to our reasoning module. 

\begin{table}[ht]

\resizebox{0.45\textwidth}{!}{
\begin{tabular}{@{}l|ccc|ccc|ccc@{}}
\toprule
          & \multicolumn{3}{c|}{Task 1}                  & \multicolumn{3}{c|}{Task 2} & \multicolumn{3}{c}{Task 3} \\ \midrule
Team      & db3  & md\_dh & {\ul \textbf{ElectricSheep}} & db3     & APEX   & md\_dh   & db3   & APEX  & vslyu-team \\
Score(\%) & 28.4 & 24.0   & 21.8                         & 42.7    & 41.0   & 31.0     & 47.8  & 44.9  & 25.6       \\ \bottomrule
\end{tabular}
}

\caption{The evaluation results in the private test set.}
\label{tab:private-main}
\vspace{-0.5cm}
\end{table}

\begin{figure}[ht]
    \centering
    \includegraphics[width=0.95\linewidth]{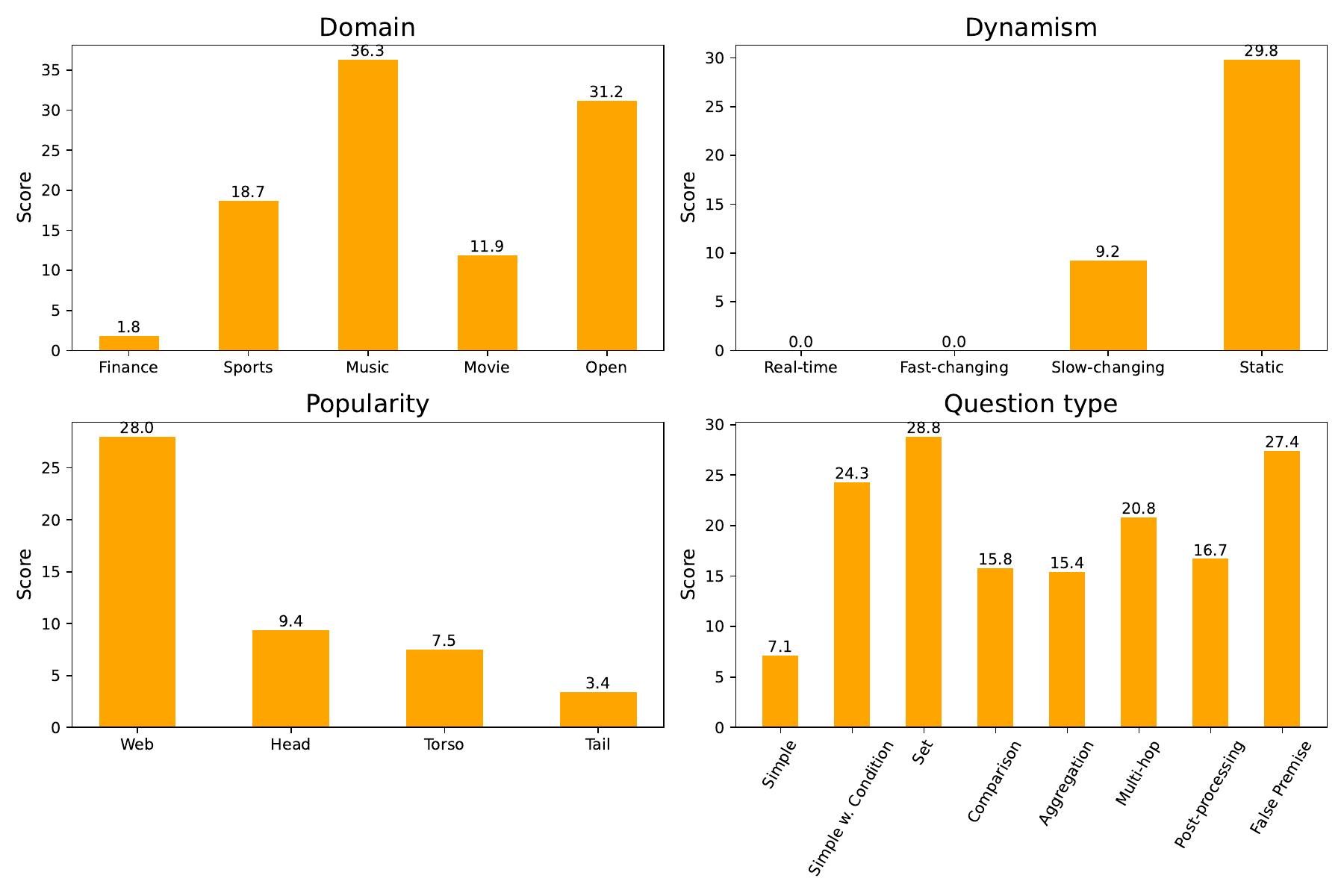}
    \caption{Detailed score across different attributes in the online private evaluation Task 1.}
    \label{fig:detailed-task-1}
\end{figure}

Furthermore, Figure~\ref{fig:detailed-task-1} illustrates the scores across various attributes, revealing that our focus has been on static and slow-changing questions, neglecting the challenging time-varying ones. This deficiency in handling dynamic questions has also led to suboptimal performance in financial question types. The score results align with those from our internal assessment; however, we encountered a higher number of incorrect responses in the movie domain and on simple question types. We hypothesize that this discrepancy may stem from variations in question distribution between our local evaluation and online test datasets. Additionally, our performance in popularity aligns well with the expectations set forth in the CRAG Benchmark \cite{yang2024cragcomprehensiverag}.

\begin{table}[ht]

\resizebox{0.35\textwidth}{!}{
\begin{tabular}{@{}l|ll@{}}
\toprule
Question Type        & Team                         & Score(\%) \\ \midrule
simple\_w\_condition & {\ul \textbf{ElectricSheep}} & 23.9      \\
set                  & {\ul \textbf{ElectricSheep}} & 36.65     \\
comparison           & dRAGonRAnGers                & 38        \\
aggregation          & {\ul \textbf{ElectricSheep}} & 18.75     \\
multi\_hop           & {\ul \textbf{ElectricSheep}} & 23.2      \\
post\_processing     & {\ul \textbf{ElectricSheep}} & 11.75     \\
false\_premise       & Future                       & 64.6      \\ \bottomrule
\end{tabular}
}

\caption{The scores of the prizes we won in Task 2.}
\label{tab:task-2-prizes}
\vspace{-0.5cm}
\end{table}

\section{Conclusion}\label{sec:conclusion}
With the designed RAG system, we finally got 3rd place in task 1 and got the prize for 5 out of 7 question types in task 2. The final evaluation scores can be found in the {\color{magenta}\href{https://discourse.aicrowd.com/t/meta-crag-challenge-2024-winners-announcement/10786}{Winners Announcement}}. The table extractor, reasoning module, and calculator module have demonstrated substantial enhancements over the baseline system. Additionally, the ICL attribute predictor has significantly reduced hallucinations in responses to difficult questions.

\newpage
\section*{Discussions}\label{sec:discuss}
Many aspects of our system can be improved. For example, we only used two tower models for retrieval, while re-ranker models are more suitable for task 1. Moreover, a two-stage retrieval and re-ranker system should be used for task 3. We didn't optimize the KG information retrieval in task 2, which can be improved a lot in the future. The current methods for handling tables are relatively simple. Some tables are useless or too large with a lot of noise, but we didn't handle these cases. There should be retrieval and structural query methods specifically designed for tables.

\begin{acks}
This project is partially supported by the National Key Research and Development Program of China with Grant No. 2023YFC3341203 as well as the National Natural Science Foundation of China with Grant Number 62276002.

We thank Prof. Ming Zhang's mentorship and support to the team and Alan Wu's assistance and discussions.
\end{acks}

\newpage
\bibliographystyle{ACM-Reference-Format-num}
\bibliography{sample-base,custom}

\newpage
\appendix

\section{Hyper-Parameters of the Final System}
We list the hyper-parameters of our final system here in Table~\ref{tab:hyper}, more details can be found in our {\color{magenta}\href{https://gitlab.aicrowd.com/shizueyy/crag-new/-/blob/main/models/model_task_1_add_direct_answer.py?ref_type=heads#L495}{source code}}. Moreover, we enable all modules for the static questions.

\begin{table}[h]
\begin{tabular}{@{}lc@{}}
\toprule
Hyper-Parameter  & Value                          \\ \midrule
Chunk Size       & $200\times 3$                  \\
Chunk Number     & $10$                           \\
Embedding Model  & sentence-t5-large              \\
Retrieval Method & two tower \& cosine similarity \\
Max Table Length & $4000$                         \\ \bottomrule
\end{tabular}
\caption{Hyper-parameters of our final system.}
\label{tab:hyper}
\end{table}

\section{Question Identification}
\label{app:ques-classify}
The source code we used to identify questions is shown as follows:

\lstset{
 basicstyle=\tiny\ttfamily,
 breaklines,
 columns=fixed,       
 numbers=left,                                        
 numberstyle=\tiny\color{gray},                            
 numbersep=5pt,
 xleftmargin=5pt,
 frame=single,
 backgroundcolor=\color[RGB]{245,245,244},            
 keywordstyle=\color[RGB]{40,40,255},                 
 commentstyle=\color[RGB]{0,96,96},                
 stringstyle=\color[RGB]{128,0,0},   
 showstringspaces=false,                              
 emph={sentence,START\_WORDS}, 
 emphstyle=\color{RubineRed}, 
 columns=flexible,
}

\begin{spacing}{1}
\begin{lstlisting}[language={Python}]
START_WORDS = [
    "who", "what", "when", "where", "why",
    "how", "is", "can", "does", "do", "did",
    "will", "would", "could", "should",
    "are", "was", "were",
    "has", "have", "had",
    "which", "whom", "whose",
]
def is_question(sentence: str):
    return sentence.lower().endswith("?") or \
           sentence.lower().startswith(tuple(START_WORDS))
\end{lstlisting}
\end{spacing}

\section{Table Transformation}
\label{app:table-transformation}
The code we used for table transformation is shown as follows.
\begin{spacing}{1}
\begin{lstlisting}[language={Python}]
from bs4 import BeautifulSoup
def clean_html(webpage, remove_links=False):
    soup = BeautifulSoup(webpage, features="html.parser")
    for script in soup.find_all(["script", "style", "meta"]):
        script.decompose()
    if remove_links:
        for link in soup.find_all("a"):
            link.unwrap()
    for footer in soup.find_all("footer"):
        footer.decompose()
    for button in soup.find_all("button"):
        button.decompose()
    for text_element in soup.find_all(String=True):
        if "\\u" in text_element or "\\U" in text_element:
            # Decode the escaped unicode characters
            decoded_text = bytes(text_element, "utf-8").decode("unicode_escape")
            text_element.replace_with(decoded_text)
    return soup
def check_empty_table(table_str: str) -> bool:
    table_str = (
        table_str.replace("|", "")
        .replace("-", "")
        .replace(" ", "")
        .replace("\n", "")
        .strip()
    )
    table_str = "".join(table_str.split())
    if table_str == "":
        return True
    return False
def get_tables(soup, html_name: str) -> list[str]:
    markdown_tables = []
    # Find all table tags
    tables = soup.find_all("table")
    for table in tables:
        markdown_table = []
        rows = table.find_all("tr")
        for row in rows:
            cells = row.find_all(["th", "td"])
            # Extract text from each cell and join them with a pipe
            formatted_row = (
                "| " + " | ".join(cell.get_text(strip=True) for cell in cells) + " |"
            )
            markdown_table.append(formatted_row)
        # Add a separator after the header (assumes the first row is header)
        if markdown_table:
            header_sep = (
                "| "
                + " | ".join("---" for _ in markdown_table[0].split("|")[1:-1])
                + " |"
            )
            markdown_table.insert(1, header_sep)
        markdown_table_str = "\n".join(markdown_table)
        if check_empty_table(markdown_table_str):
            continue
        markdown_table_str = f"Page name: {html_name}\n{markdown_table_str}"
        markdown_tables.append(markdown_table_str)
    return markdown_tables
\end{lstlisting}
\end{spacing}

\section{Prompt Details}
\subsection{Prompt for Attribute Predictor}
\label{app:prompt-attr-predictor}
Here we show the prompt we used for the attribute predictor. Specifically, we show the prompt template for dynamic prediction.
\lstset{
 basicstyle=\tiny\ttfamily,
 breaklines,
 columns=fixed,       
 numbers=left,                                        
 numberstyle=\tiny\color{gray},                            
 numbersep=5pt,
 xleftmargin=5pt,
 frame=single,
 backgroundcolor=\color[RGB]{245,245,244},            
 keywordstyle=\color[RGB]{40,40,255},                 
 commentstyle=\color[RGB]{0,96,96},                
 stringstyle=\color[RGB]{128,0,0},   
 showstringspaces=false,                              
 emph={system\_prompt, user\_prompt,few\_shot\_num}, 
 emphstyle=\color{RubineRed}, 
 columns=flexible,
}
\begin{spacing}{1}
\begin{lstlisting}[language={Python}]
{
"system_prompt": "You will be provided with a question. Your task is to identify whether this question is a static question or a dynamic question. A static question is that the answer is fixed and will not change over time. A dynamic question is that the answer will change over time or needs time information. You **MUST** choose from one of the following choices: [\"static\", \"dynamic\"]. You **MUST** give the question type succinctly, using the fewest words possible.\nHere are some examples:\n" + \
    "------\n### Question: {}\n### Static or Dynamic: {}\n\n".format(
        example["query"],
        example["static_or_dynamic"]
    ) * few_shot_num
"user_prompt": "Here is the question: {query}\nRemember your rule: You **MUST** choose from the following choices: [\"static\", \"dynamic\"].\nWhat is the static or dynamic of this question?".format(query=query)
}
\end{lstlisting}
\end{spacing}
\subsection{Prompt for Numerical Calculator}
\label{app:prompt-calculator}
Here we show the prompt template for the numerical calculator.
\lstset{
 basicstyle=\tiny\ttfamily,
 breaklines,
 columns=fixed,       
 numbers=left,                                        
 numberstyle=\tiny\color{gray},                            
 numbersep=5pt,
 xleftmargin=5pt,
 frame=single,
 backgroundcolor=\color[RGB]{245,245,244},            
 keywordstyle=\color[RGB]{40,40,255},                 
 commentstyle=\color[RGB]{0,96,96},                
 stringstyle=\color[RGB]{128,0,0},   
 showstringspaces=false,                              
 emph={}, 
 emphstyle=\color{RubineRed}, 
 columns=flexible,
}
\begin{spacing}{1}
\begin{lstlisting}[language={Python}]
class CalcAgent:
    def __init__(self, max_table_length=6000):
        self.max_table_length = max_table_length
        self.expression_sample_num = 5

    def format_expression_prompts(
        self,
        batch_queries: list[str],
        batch_retrieval_results: list[list[str]],
        batch_tables: list[list[str]],
        tokenizer: AutoTokenizer,
    ) -> list[str]:
        system_prompt = """You are provided with a question and various references. Your task is to generate a possible useful expression that is needed to answer the question. Here are the rules:
1. The expression **MUST** be a valid Python expression.
2. The expression **MUST** be useful to answer the question.
3. If you think no expression is needed, you **MUST** answer with empty string.
4. The output should be succinct, you **MUST** do reasoning in your heart without outputing the reasoning.
5. You **MUST NOT** output any other words except the valid Python expression.
6. You **MUST NOT** output the expression that need the user to input anything.
"""
        formatted_prompts = []
        for _idx, query in enumerate(batch_queries):
            retrieval_results = batch_retrieval_results[_idx]
            related_tables = batch_tables[_idx]
            user_message = ""
            references = ""
            if len(retrieval_results) > 0:
                references += "# References \n"
                # Format the top sentences as references in the model's prompt template.
                for _snippet_idx, snippet in enumerate(retrieval_results):
                    references += f"- {snippet.strip()}\n"
            user_message += f"{references}\n------\n\n"

            if len(related_tables) > 0:
                table_references = ""
                user_message += "## Table references \n"
                for idx, table in enumerate(related_tables):
                    table_references += f"### Table {idx + 1}: \n"
                    table_references += f"{table}\n"
                table_references = table_references[: self.max_table_length]
                user_message += f"{table_references}\n------\n\n"
            user_message += "**Remember your rules**:"
            user_message += """
1. The expression **MUST** be a valid Python expression.
2. The expression **MUST** be useful to answer the question.
3. If you think no expression is needed, you **MUST** answer with empty string.
4. The output should be succinct, you **MUST** do reasoning in your heart without outputing the reasoning.
5. You **MUST NOT** output any other words except the valid Python expression.
6. You **MUST NOT** output the expression that need the user to input anything.
"""
            user_message += f"Question: {query}\n"
            user_message += f"Using the references listed above and based on the question, generate a valid Python expression for me: \n"

            formatted_prompts.append(
                tokenizer.apply_chat_template(
                    [
                        {"role": "system", "content": system_prompt},
                        {"role": "user", "content": user_message},
                    ],
                    tokenize=False,
                    add_generation_prompt=True,
                )
            )
        return formatted_prompts
\end{lstlisting}
\end{spacing}

\subsection{Prompt for LLM Knowledge Extractor}
\label{app:prompt-knowledge-extractor}
Here is the prompt template for the LLM Knowledge Extractor.
\lstset{
 basicstyle=\tiny\ttfamily,
 breaklines,
 columns=fixed,       
 numbers=left,                                        
 numberstyle=\tiny\color{gray},                            
 numbersep=5pt,
 xleftmargin=5pt,
 frame=single,
 backgroundcolor=\color[RGB]{245,245,244},            
 keywordstyle=\color[RGB]{40,40,255},                 
 commentstyle=\color[RGB]{0,96,96},                
 stringstyle=\color[RGB]{128,0,0},   
 showstringspaces=false,                              
 emph={self,system\_prompt,user\_message,formatted\_prompts,queries}, 
 emphstyle=\color{RubineRed}, 
 columns=flexible,
}
\begin{spacing}{1}
\begin{lstlisting}[language={Python}]
class CragSystem:
    def get_direct_answers(self, queries) -> list[str]:
        system_prompt = """You are provided with a question.
Your task is to answer the question with your reasoning process.
If you can't answer it directly based on your knowledge, respond with 'I don't know'.
If you think the premise of the question is wrong, for example, the question asks information about a person's husband, but you are sure that the person doesn't have one, you should answer with "Invalid question" without any other words.
You **MUST** think if the question has a false premise, then think the final answer.
You **MUST** generate the reasoning process before the answer. You **MUST** generate your output with the following format:

===START===
## Reasoning:
- Does it have a false premise?
<YOUR REASONING>
- What is the final answer?
<YOUR REASONING>
------
## Answer:
<YOUR FINAL ANSWER>
===END===

**IMPORTANT RULES**:
- If you can't answer it directly based on your knowledge, respond with 'I don't know'.
- Your generation **MUST** starts with "===START===" and ends with "===END===".
- `<YOUR FINAL ANSWER>` should be succinct, and use as few words as possible.
- `<YOUR REASONING>` should be a detailed reasoning process that explains how you arrived at your answer.
- If you think the premise of the question is wrong, for example, the question asks information about a person's husband, but you are sure that the person doesn't have one, you should answer with "Invalid question" without any other words.
Let's think step by step now!"""
        formatted_prompts = []
        for _idx, query in enumerate(queries):
            user_message = query
            formatted_prompts.append(
                self.tokenizer.apply_chat_template(
                    [
                        {"role": "system", "content": system_prompt},
                        {"role": "user", "content": user_message},
                    ],
                    tokenize=False,
                    add_generation_prompt=True,
                )
            )
\end{lstlisting}
\end{spacing}

\subsection{Prompt for KG Module}
\label{app:prompt-kg}
We show the prompt template of the function-calling method in the knowledge graph module.

\lstset{
 basicstyle=\tiny\ttfamily,
 breaklines,
 columns=fixed,       
 numbers=left,                                        
 numberstyle=\tiny\color{gray},                            
 numbersep=5pt,
 xleftmargin=5pt,
 frame=single,
 backgroundcolor=\color[RGB]{245,245,244},            
 keywordstyle=\color[RGB]{40,40,255},                 
 commentstyle=\color[RGB]{0,96,96},                
 stringstyle=\color[RGB]{128,0,0},   
 showstringspaces=false,                              
 emph={self,system\_prompt,user\_message,formatted\_prompts,queries,query\_times}, 
 emphstyle=\color{RubineRed}, 
 columns=flexible,
}
\begin{spacing}{1}
\begin{lstlisting}[language={Python}]
class KGToolRAGModel
    def get_tool_references(
        self, queries: list[str], query_times: list[str]
    ) -> list[list[str]]:
        system_prompt = f"""
You are a helpful assistant in function calling. I have a knowledge graph and a set of functions that can be called. You will be given a question and the query time. Your task is to generate several function calls that can help me answer the question. Here are functions and their descriptions:
{TOOLS}
Remember your rules:
1. You **MUST** follow the function signature.
2. You **MUST** output the JSON format that can be read by `json.loads`. Return empty list if no useful function calls can be found.
3. For each function call, you should output its function name and corresponding arguments.

Here are examples:

# Example 1:
Query: which company have larger market cap, hri or imppp?
Query time: 03/13/2024, 10:19:56 PT
Output: [
    {{"function_name": "finance_get_market_capitalization", "args": ["hri"]}},
    {{"function_name": "finance_get_market_capitalization", "args": ["imppp"]}}
]

# Example 2:
Query: who are the current members of the band eagles?
Query time: 03/05/2024, 23:17:59 PT
Output: [
    {{"function_name": "music_get_members", "args": ["eagles"]}}
]
"""
        formatted_prompts = []
        for idx, query in enumerate(queries):
            user_message = f"Question: {query}\n"
            user_message += f"Query time: {query_times[idx]}\n"
            user_message += "Using the tools listed above and based on the question, generate useful function calls for me. \n"
            user_message += """
Remember your rules:
1. You **MUST** follow the function signature.
2. You **MUST** output the JSON format that can be read by `json.loads`.
3. For each function call, you should output its function name and corresponding arguments.
"""
            formatted_prompts.append(
                self.llm.get_tokenizer().apply_chat_template(
                    [
                        {"role": "system", "content": system_prompt},
                        {"role": "user", "content": user_message},
                    ],
                    tokenize=False,
                    add_generation_prompt=True,
                )
            )
\end{lstlisting}
\end{spacing}

\subsection{Prompt for Reasoning Module}
\label{app:prompt-reasoning}
We show the prompt template of the reasoning module, where we successfully controlled the reasoning paths and output format and many corner cases.
\lstset{
 basicstyle=\tiny\ttfamily,
 breaklines,
 columns=fixed,       
 numbers=left,                                        
 numberstyle=\tiny\color{gray},                            
 numbersep=5pt,
 xleftmargin=5pt,
 frame=single,
 backgroundcolor=\color[RGB]{245,245,244},            
 keywordstyle=\color[RGB]{40,40,255},                 
 commentstyle=\color[RGB]{0,96,96},                
 stringstyle=\color[RGB]{128,0,0},   
 showstringspaces=false,                              
 emph={self,system\_prompt,user\_message,formatted\_prompts,queries,query\_times}, 
 emphstyle=\color{RubineRed}, 
 columns=flexible,
}
\begin{spacing}{1}
\begin{lstlisting}[language={Python}]
system_prompt = """You are provided with a question and various references.
Your task is to answer the question with your reasoning process.
There are also some calculation results from another agent, which may be useful for you.
There is an answer from another agent which may be useful. It may have hallucination. You need to judge whether to trust it by yourself.
If the references do not contain the necessary information to answer the question and you can't answer it directly based on your knowledge, respond with 'I don't know'.
If you think the premise of the question is wrong, for example, the question asks information about a person's husband, but you are sure that the person doesn't have one, you should answer with "Invalid question" without any other words.
You **MUST** think if the question has a false premise, then think the final answer.
You **MUST** generate the reasoning process before the answer. You **MUST** generate your output with the following format:

===START===
## Reasoning:
- Does it have a false premise?
<YOUR REASONING>
- What is the final answer?
<YOUR REASONING>
- Can you answer it based on current knowledge?
<YOUR REASONING>
------
## Answer:
<YOUR FINAL ANSWER>
## False Premise:
<HAS_FALSE_PREMISE_OR_NOT>
===END===

**IMPORTANT RULES**:
- If the references do not contain the necessary information to answer the question and you can't answer it directly based on your knowledge, respond with 'I don't know'.
- Your generation **MUST** starts with "===START===" and ends with "===END===".
- `<YOUR FINAL ANSWER>` should be succinct, and use as few words as possible.
- `<YOUR REASONING>` should be a detailed reasoning process that explains how you arrived at your answer.
- `<HAS_FALSE_PREMISE_OR_NOT>` should be "yes" if the premise is wrong and the question is invalid, and "no" otherwise. It can **ONLY** be chosen from these two options.
- If you think the premise of the question is wrong, for example, the question asks information about a person's husband, but you are sure that the person doesn't have one, you should answer with "Invalid question" without any other words.
Let's think step by step now!"""
user_message = ""
references = ""
if len(retrieval_results) > 0:
    references += "# References \n"
    # Format the top sentences as references in the model's prompt template.
    for _snippet_idx, snippet in enumerate(retrieval_results):
        references += f"- {snippet.strip()}\n"
user_message += f"{references}\n------\n\n"
if len(kg_results) > 0:
    kg_references = ""
    user_message += "## Knowledge Graph references \n"
    for idx, kg_result in enumerate(kg_results):
        kg_references += f"### KG Ref {idx + 1}: \n"
        kg_references += f"{kg_result}\n"
    kg_references = kg_references[:1000]
    user_message += f"{kg_references}\n------\n\n"
    logger.debug("Currect KG references:\n{}".format(kg_references))
if len(related_tables) > 0:
    table_references = ""
    user_message += "## Table references \n"
    for idx, table in enumerate(related_tables):
        table_references += f"### Table {idx + 1}: \n"
        table_references += f"{table}\n"
    table_references = table_references[: self.max_table_length]
    user_message += f"{table_references}\n------\n\n"
if len(generated_expressions) > 0:
    expression_references = ""
    user_message += "## Possible useful calculation results \n"
    for idx, expression in enumerate(generated_expressions):
        expression_references += f"### Calculation {idx + 1}: \n"
        expression_references += f"{expression}\n"
    user_message += f"{expression_references}\n------\n\n"
if direct_answer is not None:
    user_message += (
        f"# An answer from another agent:\n{direct_answer}\n------\n\n"
    )
user_message += """**Remember your IMPORTANT RULES**:
- If the references do not contain the necessary information to answer the question and you can't answer it directly based on your knowledge, respond with 'I don't know'.
- Your generation **MUST** starts with "===START===" and ends with "===END===".
- `<YOUR FINAL ANSWER>` should be succinct, and use as few words as possible.
- `<YOUR REASONING>` should be a detailed reasoning process that explains how you arrived at your answer.
- `<HAS_FALSE_PREMISE_OR_NOT>` should be "yes" if the question is invalid, and "no" otherwise. It can **ONLY** be chosen from these two options.
- If you think the premise of the question is wrong, for example, the question asks information about a person's husband, but you are sure that the person doesn't have one, you should answer with "Invalid question" without any other words.
"""
user_message += (
    f"Using the references listed above, answer the following question: \n"
)
user_message += f"Current Time: {query_time}\n"
user_message += f"Question: {query}\n"
user_message += "Let's think step by step now!\n"
\end{lstlisting}
\end{spacing}
As illustrated in the prompt template, we have devised a set of rules to handle corner cases and enhance the accuracy of the model's responses. We repeat the important rules in both the system prompt and the user prompt to strengthen the control.

\subsection{Prompt for Backup Summarization Agent}
\label{app:prompt-backup}
We show the prompt template of the backup summarization agent below.
\lstset{
 basicstyle=\tiny\ttfamily,
 breaklines,
 columns=fixed,       
 numbers=left,                                        
 numberstyle=\tiny\color{gray},                            
 numbersep=5pt,
 xleftmargin=5pt,
 frame=single,
 backgroundcolor=\color[RGB]{245,245,244},            
 keywordstyle=\color[RGB]{40,40,255},                 
 commentstyle=\color[RGB]{0,96,96},                
 stringstyle=\color[RGB]{128,0,0},   
 showstringspaces=false,                              
 emph={self,system\_prompt,user\_message,formatted\_prompts,queries,reasoning\_answers}, 
 emphstyle=\color{RubineRed}, 
 columns=flexible,
}
\begin{spacing}{1}
\begin{lstlisting}[language={Python}]
class SummarizationAgent
    def format_prompts(
        self,
        queries,
        reasoning_answers: list[str] = [],
    ):
        assert len(queries) == len(reasoning_answers)
        system_prompt = "You are provided with a question and a reasoning process from another agent. Your task is to summarize the reasoning process and finally answer the question succinctly, using the fewest words possible."
        formatted_prompts = []
        for _idx, query in enumerate(queries):
            reasoning_process = reasoning_answers[_idx]
            user_message = ""
            user_message += f"Question: {query}\n"
            user_message += (
                f"# Useful Reasoning Process: \n{reasoning_process}\n-----\n\n"
            )
            user_message += f"Using the reasoning process above, answer the question."
            formatted_prompts.append(
                self.tokenizer.apply_chat_template(
                    [
                        {"role": "system", "content": system_prompt},
                        {"role": "user", "content": user_message},
                    ],
                    tokenize=False,
                    add_generation_prompt=True,
                )
            )
        return formatted_prompts
\end{lstlisting}
\end{spacing}

\end{document}